\documentclass[12pt]{article}
\usepackage{amsmath,amssymb}
\usepackage{amsmath,amsthm,mathtools}
\usepackage{graphicx,epstopdf}
\usepackage{algorithm}
\usepackage{algorithmic}
\usepackage{booktabs} 
\usepackage[dvipsnames]{xcolor}
\usepackage{float}
\usepackage{bm}
\usepackage{hyperref} 
\usepackage{lineno}
\hypersetup{
colorlinks=true,
citecolor=blue,
linkcolor=red,
anchorcolor=black,
urlcolor=blue,
pdfsubject={LaTeX},
}

\DeclareMathOperator*{\argmax}{arg\,max}

\usepackage{amsfonts,amssymb}       

\usepackage{authblk}
\usepackage[authoryear]{natbib}

\usepackage{geometry}
\usepackage{pdflscape}
\usepackage{array}
\renewcommand{\arraystretch}{2}
\usepackage{array}
\newcommand{\PreserveBackslash}[1]{\let\temp=\\#1\let\\=\temp}
\newcolumntype{C}[1]{>{\PreserveBackslash\centering}p{#1}}
\newcolumntype{R}[1]{>{\PreserveBackslash\raggedleft}p{#1}}
\newcolumntype{L}[1]{>{\PreserveBackslash\raggedright}p{#1}}
\newcommand{\red}[1]{{\color{black} #1}}

\def\x{\mathbf x}
\def\y{\mathbf y}
\def\z{\mathbf z}
\def\b{\mathbf b}
\def\r{\mathbf r}
\def\W{{\mathbf W}}
\def\L{{\mathbf L}}
\def\Tr{{\rm Tr}}
\def\R{\mathbb{R}}

\title{Contrastive Similarity Matching for Supervised Learning}

\author[1]{Shanshan Qin} 
\author[2]{Nayantara Mudur} 
\author[1]{Cengiz Pehlevan} 
\affil[1]{John A. Paulson School of Engineering and Applied Sciences, Harvard University, Cambridge, MA USA}
\affil[2]{Department of Physics, Harvard University, Cambridge, MA USA}

\date{\vspace{-5ex}}
\begin{document}

\maketitle

\begin{abstract}

We propose a novel biologically-plausible solution to the credit assignment problem motivated by observations in the ventral visual pathway and trained deep neural networks. In both, representations of objects in the same category become progressively more similar, while objects belonging to different categories become less similar. We use this observation to motivate a layer-specific learning goal in a deep network: each layer aims to learn a representational similarity matrix that interpolates between previous and later layers. We formulate this idea using a contrastive similarity matching objective function and derive from it deep neural networks with feedforward, lateral, and feedback connections, and neurons that exhibit biologically-plausible Hebbian and anti-Hebbian plasticity. Contrastive similarity matching can be interpreted as an energy-based learning algorithm, but with significant differences from others in how a contrastive function is constructed.

\end{abstract}

\section{Introduction}

Synaptic plasticity is generally accepted as the underlying mechanism of learning in the brain, which almost always involves a large population of neurons and synapses across many different brain regions. How the brain modifies and coordinates individual synapses in the face of limited information available to each synapse in order to achieve a global learning task, the credit assignment problem, has puzzled scientists for decades. A major effort in this domain has been to look for a biologically-plausible implementation of the back-propagation of error algorithm (BP) \citep{rumelhart1986learning}, which has long been disputed due to its biological implausibility \citep{crick1989recent}, although recent studies have made progress in resolving some of these concerns \citep{xie2003equivalence,lee2015difference,lillicrap2016random,nokland2016direct,scellier2017equilibrium,guerguiev2017towards,whittington2017approximation,sacramento2018dendritic,richards2019dendritic,whittington2019theories,belilovsky2018greedy,ororbia2019biologically,lillicrap2020backpropagation}.

\begin{figure}[t]
\centering
\includegraphics[width=0.5\linewidth]{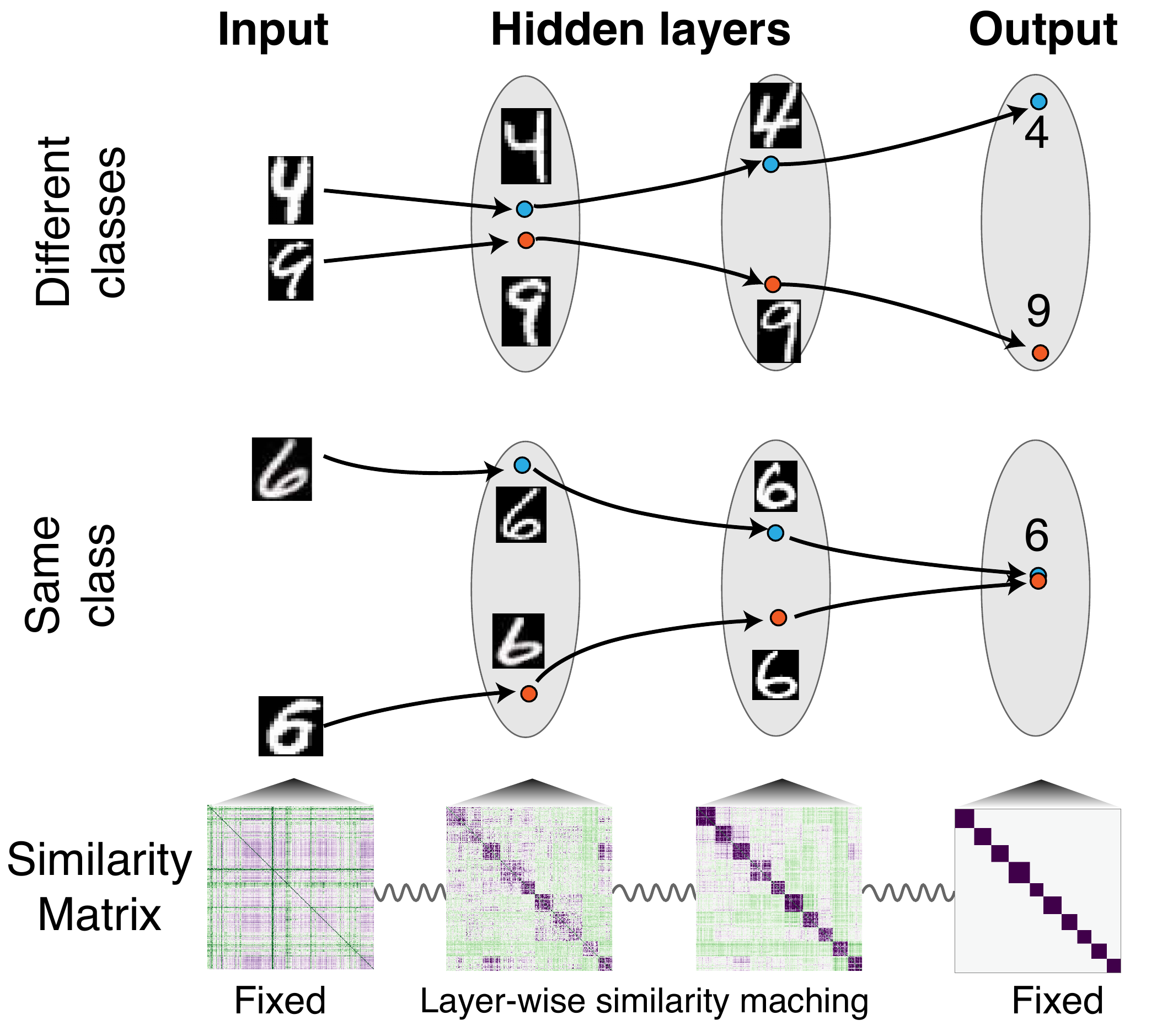}
\caption{Supervised learning via layer-wise similarity matching. For inputs of different categories, similarity-matching differentiates the representations progressively (top), while for objects of the same category, representations become more and more similar (middle). For a given set of training data and their corresponding labels, the training process can be regarded as learning hidden representations whose similarity matrices match that of both input and output (bottom). The tuning of representational similarity is indicated by the springs with the constraints that input and output similarity matrices are fixed.}
\label{fig:supervised_illustr_2}
\end{figure}

In this paper, we present a novel approach to the credit assignment problem, motivated by observations on the nature of hidden layer representations in the ventral visual pathway of the brain and deep neural networks. In both, representations of objects belonging to different categories become less similar, while representations of objects belonging to the same category become more similar \citep{grill2014functional,kriegeskorte2008matching,yamins2016using}. In other words, categorical clustering of representations becomes more and more explicit in the later layers (Fig. \ref{fig:supervised_illustr_2}). These results suggest a new approach to the credit assignment problem.  By assigning each layer a layer-local similarity matching task \citep{pehlevan2019neuroscience,obeid2019structured}, whose goal is to learn an intermediate representational similarity matrix between previous and later layers, we may be able to get away from the need of backward propagation of errors (Fig. \ref{fig:supervised_illustr_2}).  Motivated by this idea and previous observations that error signal can be implicitly propagated via the change of neural activities \citep{hinton1988learning,scellier2017equilibrium}, we propose \red{a biologically plausible supervised learning algorithm, the contrastive similarity matching (CSM) algorithm.} 

\red{Contrastive training \citep{anderson1987mean,movellan1991contrastive,baldi1991contrastive} has been used to learn the energy landscapes of neural networks (NNs) whose dynamics minimize an energy function. Examples include influential algorithms like the Contrastive Hebbian Learning (CHL) \citep{movellan1991contrastive} and Equilibrium Propagation (EP) \citep{scellier2017equilibrium}, where weight-updates rely on the difference of the neural activity between a free phase, and a clamped (CHL) or nudged (EP) phase to locally approximate the gradient of an error signal. The learning process can be interpreted as minimizing a contrastive function, which reshapes the energy landscape to eliminate spurious fixed points and makes the desired fixed point more stable. 

The CSM algorithm applies this idea to a contrastive function formulated by nudging the output neurons of a multilayer similarity matching objective function \citep{obeid2019structured}. As a consequence, the hidden layers learn intermediate representations between their previous and later layers. From the CSM contrastive function, we derive deep neural networks  with feedforward, lateral and feedback connections, and neurons that exhibit biologically-plausible Hebbian and anti-Hebbian plasticity. 

The nudged phase of the CSM algorithm is analogous to the nudged phase of EP but different. It performs Hebbian feedforward and anti-Hebbian lateral updates. CSM  has opposite sign for the lateral connection updates compared with EP and CHL. This is because our weight updates solve a minimax problem. Anti-Hebbian learning pushes neurons within a layer to learn different representations. The free phase of CSM is also different where only feedforward weights are updated by an anti-Hebbian rule. In EP and CHL all weights are updated.}

Our main contributions and results are listed below:
\begin{itemize}
    \item \red{We provide a novel approach to the credit assignment problem using biologically-plausible learning rules by generalizing the similarity matching principle \citep{pehlevan2019neuroscience} to supervised learning tasks and introducing the Contrastive Similarity Matching algorithm. }
    
    \item The proposed supervised learning algorithm can be related to other energy-based algorithms, but with a distinct underlying mechanism.
    
    \item We present a version of our neural network algorithm with structured connectivity.
    
    \item We show that the performance of our algorithm is on par with other energy-based algorithms using numerical simulations. The learned representations of our Hebbian/anti-Hebbian network is sparser. 
\end{itemize}

The rest of this paper is organized as follows. \red{In Section \ref{sec:csm}, to illustrate our main ideas we introduce and discuss supervised similarity matching. We then introduce} nudged deep similarity matching objective, from which we derive the CSM algorithm for deep neural networks with nonlinear activation functions and structured connectivity. We discuss the relation of CSM to other energy-based learning algorithms. In Section \ref{sec:simulations}, we report the performance of CSM and compare it with EP, highlighting the differences between them. Finally, we discuss our results, possible biological mechanisms, and relate them to other works in Section \ref{sec:discussion}.

\section{Contrastive Similarity Matching for Deep Nonlinear Networks}\label{sec:csm}

\subsection{Warm Up: Supervised Similarity Matching Objective}



Here we illustrate our main idea in a simple setting. Let $\x_t \in \R^n$, $t= 1,\cdots,T$ be a set of data points and  
$\z_t^l \in \R^k$ be their corresponding desired output or labels. 
Our idea is that the representation learned by the hidden layer, $\y_t \in \R^m$, should be half-way between the input $\x$ and the desired output $\z^l$. We formulate this idea using representational similarities, quantified by the dot product of representational vectors within a layer. Our proposal can be formulated as the following optimization problem, which we name supervised similarity matching:
\begin{linenomath*}
\begin{equation}\label{eq:targetFunLinear}
    \min _{\lbrace \y_t\rbrace_{t=1}^T}\frac{1}{T^2}\sum_{t=1}^T\sum_{t'=1}^T[(\x_t^{\top}\x_{t'} - \y_t^{\top}\y_{t'})^2 + (\y_t^{\top}\y_{t'} - \z_t^{l\top}\z_{t'}^l)^2].
\end{equation}
\end{linenomath*}

To get an intuition about what this cost function achieves, consider the case where only one training datum exists. Then, $\y_1^{\top}\y_{1} = \frac{1}{2}(\x_1^{\top}\x_{1} + \z_1^{l\top}\z_{1}^l)$, satisfying our condition. When multiple training data are involved, interactions between different data points lead to a non-trivial solution, but the fact that the hidden layer representations are in between the input and output layers stays.   

The optimization problem \eqref{eq:targetFunLinear} can be analytically solved, making our intuition precise. Let the representational similarity matrix of the input layer be $R_{tt'}^x\equiv \x_t^\top \x_{t'}$, the hidden layer be $R_{tt'}^y\equiv \y_t^\top \y_{t'}$, and the output layer be $R_{tt'}^z\equiv \z^{l\top}_t \z^l_{t'}$. Instead of solving $\y$ directly, we can reformulate and solve the supervised similarity matching problem \eqref{eq:targetFunLinear} for $\mathbf{R}^y$, and then obtain $\y$s by a matrix factorization through an eigenvalue decomposition. By completing the square, problem \eqref{eq:targetFunLinear} becomes an optimization problem for $\mathbf{R}^y$:  
\begin{linenomath*}
\begin{align}
\min_{\mathbf{R}^y\in \mathcal{S}^m} \frac 1{T^2}\left\Vert \frac 12\left(\mathbf{R}^x+\mathbf{R}^z\right) -\mathbf{R}^y\right\Vert_F^2,
\end{align}
\end{linenomath*}
where $\mathcal{S}^m$ is the set of symmetric matrices with rank $m$, and $F$ denotes the Frobenious norm. Optimal $\mathbf{R}^y$ is given by keeping the top $m$ modes in the eigenvalue decomposition of $\frac 12\left(\mathbf{R}^x+\mathbf{R}^z\right)$ and setting the rest to zero. If $m \geq {\rm rank}(\mathbf{R}^x+\mathbf{R}^z)$, then optimal $\mathbf{R}^y$ exactly equals $\frac 12\left(\mathbf{R}^x+\mathbf{R}^z\right)$, achieving a representational similarity matrix that is the average of input and output layers.

\red{The supervised similarity matching problem \eqref{eq:targetFunLinear} can be solved by an online algorithm that can in turn be mapped onto the operation of a biologically plausible network with a single hidden layer, which runs an attractor dynamics minimizing an energy function (see Appendix \ref{sec:SSM} for details). This approach can be generalized to multi-layer and nonlinear networks. We do not pursue it further because the resulting algorithm does not perform as well due to spurious fixed points of nonlinear dynamics for a given input $\x_t$.  The Contrastive Similarity Matching algorithm overcomes this problem.} 

\subsection{Nudged Deep Similarity Matching Objective and Its Dual Formulation}
\red{Our goal is to combine the ideas of supervised similarity matching and contrastive learning to derive a biologically plausible supervised learning algorithm. To do so, we define the nudged similarity matching problem first.}

In energy-based learning algorithms like CHL and EP, weight-updates rely on the difference of neural activity between a free phase and a clamped/nudged phase to locally approximate the gradient of an error signal. This process can be interpreted as minimizing a contrastive function, which reshapes the energy landscape to eliminate the spurious fixed points and make the fixed point corresponding to the desired output more stable. \red{We adopt this idea to introduce what we call the nudged similarity matching cost function, and derive its dual formulation, which will be the energy function used in our contrastive formulation.} 

We consider a $P$-layer ($P-1$ hidden layers) NN with nonlinear activation functions, $f$. For notational convenience, we denote inputs to the network by $\mathbf{r}^{(0)}$, outputs by $\mathbf{r}^{(P)}$, and activities of hidden layers by $\mathbf{r}^{(p)}, p = 1, \cdots, P-1$. We propose the following objective function for the training phase where outputs are nudged toward the desired labels $\z_t^l$
\begin{linenomath*}
\begin{align}{\label{eq:SM}}
    &\min_{\substack{a_1\leq\mathbf{r}_t^{p}\leq a_2 \\ t = 1,\cdots, T \\ p = 1,\cdots,P}}\sum_{p=1}^P\frac{\gamma^{p-P}}{2T^2}\sum_{t=1}^T\sum_{t'=1}^T||\mathbf{r}_t^{(p-1)\top}\mathbf{r}_{t'}^{(p-1)} - \mathbf{r}_t^{(p)\top}\mathbf{r}_{t'}^{(p)}||_2^2 \nonumber + \sum_{p = 1}^P\frac{2\gamma^{p-P}}{T}\sum_{t=1}^{T}\mathbf{F}(\mathbf{r}_t^{(p)})^{\top}\mathbf{1} \\
    &+ \frac{2\beta}{T} \sum_{t=1}^T \left \Vert \mathbf{r}^{(P)}_t - \z^l_t\right\Vert_2^2.
\end{align}
\end{linenomath*}

Here, $\beta$ is a control parameter that specifies how strong the nudge is. $\beta \to \infty$ limit corresponds to clamping the output layer to the desired output. $\gamma \geq 0$ is a parameter that controls the influence of the later layers to the previous layers. $F(\r_t^{(p)})$ is a regularizer 
defined and related to the activation function by $d F(\r_t^{(p)})/d\r_t^{(p)} = \bm{u}^{(p)}_t - \mathbf{b}_t^{(p)}$,
where $\r_t^{(p)}= f(\bm{u}^{(p)}_t)$, $\bm{u}^{(p)}_t$  and $\r_t^{(p)}$ are the total input and output of $p$-th layer respectively, $\mathbf{b}_t^{(p)}$ is the the threshold of neurons in layer $p$. The reason for the inclusion of this regularizer will be apparent below. We assume $f$ to be a monotonic and bounded function, whose bounds are given by $a_1$ and $a_2$.

The objective function \eqref{eq:SM} is almost identical to the deep similarity matching objective introduced in \citep{obeid2019structured}, except the nudging term. \citet{obeid2019structured} used $\beta = 0$ version as an unsupervised algorithm. Here, we use a non-zero $\beta$ for supervised learning.

\red{We note that we have not made a reference to a particular neural network yet. This is because the neural network that optimizes \eqref{eq:SM} will be fully derived from the nudged similarity matching problem. It will not be prescribed as in traditional approaches to deep learning. We next describe how to do this derivation.}

Using the duality transforms introduced in \citep{pehlevan2018similarity,obeid2019structured}, the above nudged supervised deep similarity matching problem \eqref{eq:SM} can be turned into a dual minimax problem:
\begin{linenomath*}
\begin{align}\label{multilayer}
\min_{\{\W^{(p)}\}}\max_{\{\L^{(p)}\}} &\frac 1T \sum_{t=1}^T l_t\left(\{\W^{(p)}\},\{\L^{(p)}\},\r^{(0)}_t,\z^l_t,\beta\right),
\end{align}
\end{linenomath*}

where
\begin{linenomath*}
\begin{align}\label{multil}
&l_t : = \min_{\substack{ a_1\leq \r_t^{(p)}  \leq a_2 \\ p = 1,\ldots,P}} \sum_{p=1}^P \gamma^{p-P} \left[{\rm Tr} {\W^{(p)}}^{\top} {\W^{(p)}} -2\r_t^{(p)\top}\W^{(p)}\r_t^{(p-1)}\right. \nonumber \\
&\left.+ \frac {1+\gamma(1- \delta_{pP})}{2}c^{(p)}\left(2\r_t^{(p)\top}\L^{(p)}\r_t^{(p)}-\rm Tr {\L^{(p)\top}}{\L^{(p)}}\right ) + 2 \mathbf{F}\left(\r_t^{(p)}\right)^{\top}\mathbf{1}\right] + 2\beta \left \Vert \r^{(P)}_t - \z^l_t\right\Vert_2^2,
\end{align}
\end{linenomath*}
Here, we introduced $c^{(p)}$ as a  parameter that governs the relative importance of forward versus recurrent inputs and $c^{(p)}=1$ corresponds to the exact transformation, details of which is given in the Appendix \ref{sec:deep_SSM}.

\begin{figure}
\centering
\includegraphics[width=0.6\linewidth]{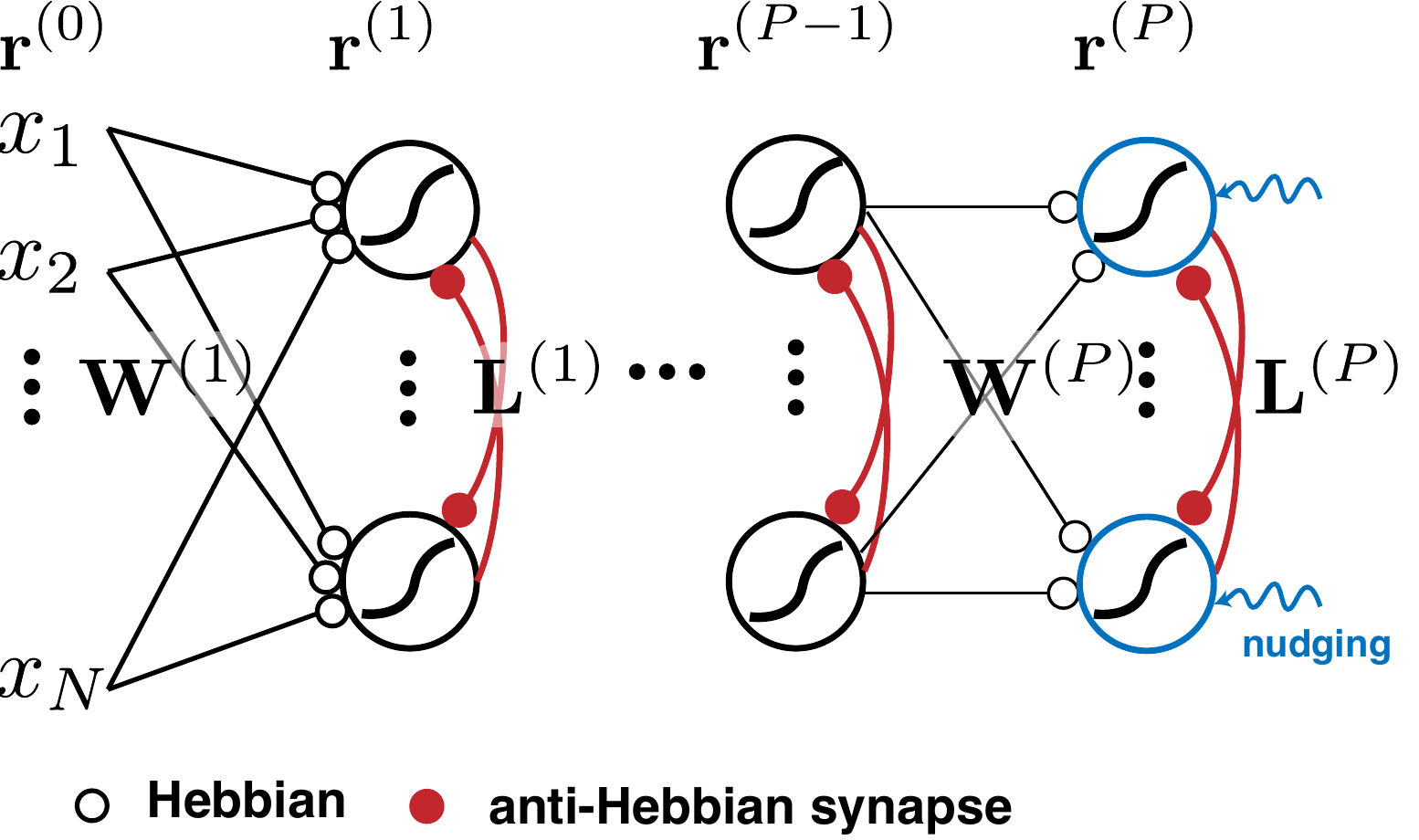}
\caption{Illustration of the Hebbian/anti-Hebbian network with $P$ hidden layers that implements the contrastive similarity matching algorithm. The output layer neurons alternate between the free phase and nudged phase.}
\label{fig:csmNet}
\end{figure}

 In Appendix \ref{sec: appen_B}, we show that the the objective of the $\min$ in $l_t$ defines an energy function for a deep neural network with feedforward, lateral and feedback connections (Figure \ref{fig:csmNet}). It has following neural dynamics:
\begin{linenomath*}
\begin{align}\label{eq:multi_layer_dyn}
   \tau_p\frac{d\bm{u}^{(p)}}{dt} &= -\bm{u}^{(p)} + \W^{(p)}\r_t^{(p-1)} - c^{(p)}[1+\gamma (1-\delta_{pP})]\L^{(p)}\r_t^{(p)} + \mathbf{b}_t^{(p)}  \nonumber \\
   & + \gamma(1-\delta_{pP})\W^{(p+1)\top}\r_t^{(p+1)} - 2\beta \delta_{pP}(\r_t^{(P)} - \z_t^l),  \nonumber \\
   \r_t^{(p)} &= \bm{f}(\bm{u}^{(p)}),
\end{align}
\end{linenomath*}
where $\delta_{pP}$ is the Kronecker delta, $p=1,\cdots,P$,  $\tau_p$ is a time constant, $\W^{(P+1)} = \bm{0}$, $\r_t^{(P+1)} = \bm{0}$.  
Therefore, the minimization can be performed by running the dynamics until convergence. This observation will be the building block of our CSM algorithm, which we present below. \red{Finally, we note that the introduction of the regularizer in \eqref{eq:SM} is necessary for the energy interpretation and for proving the convergence of the neural dynamics \citep{obeid2019structured}.}

\subsection{Contrastive Similarity Matching}

We first state our contrastive function and then discuss its implications. We suppress the dependence on training data in $l_t$ and define:
\begin{linenomath*}
\begin{align}\label{lstar}
   \{\L^{*(p)}\} \equiv \argmax_{\{\L^{(p)}\}} \frac 1T \sum_{t=1}^T l_t\left(\{\W^{(p)}\},\{\L^{(p)}\},\{\mathbf{b}^{(p)}\},\beta\right),
\end{align}
\end{linenomath*}
and
\begin{linenomath*}
\begin{equation}\label{eq:energy}
    E(\{\W^{(p)}\},\{\mathbf{b}^{(p)}\},\beta) = \frac 1T \sum_{t=1}^T l_t\left(\{\W^{(p)}\},\{\L^{*(p)}\},\{\mathbf{b}^{(p)}\},\beta\right).
\end{equation}
\end{linenomath*}
Finally, we formulate our contrastive function as
\begin{linenomath}
\begin{align}\label{cf}
 J^{\beta}(\{\W^{(p)}\},\{\mathbf{b}^{(p)}\}) = E(\beta) - E(0),
\end{align}
\end{linenomath}
which is to be minimized over feedforward and feedback weights $\{\W^{(p)}\}$, as well as bias $\{\mathbf{b}^{(p)}\}$. For fixed bias, minimization of the first term, $E(\beta)$ corresponds exactly to the optimization of the minimax dual of nudged deep similarity matching \eqref{multilayer}. The second term $E(0)$ corresponds to a free phase, where no nudging is applied. We note that in order to arrive at a contrastive minimization problem, we use the same optimal lateral weights, \eqref{lstar}, from the nudged phase in the free phase. Compared to the minimax dual of nudged deep similarity matching \eqref{multilayer}, we also optimize it over the bias for better performance.    

Minimization of the contrastive function \eqref{cf} closes the energy gap between nudged and free phases. Because the energy functions are evaluated at the fixed point of the neural dynamics \eqref{eq:multi_layer_dyn}, such procedure enforces the output of the nudged network to be a fixed point of the free neural dynamics.

To optimize our contrastive function \eqref{cf} in a stochastic (one training datum at a time) manner, we use the following procedure. For each pair of training data $\{\r_t^0, \z_t^l\}$, we run the nudged phase ($\beta \ne 0$) dynamics \eqref{eq:multi_layer_dyn} until convergence to get the fixed point $\r_{\beta,t}^{(p)}$. Next, we run the free phase ($\beta = 0$) neural dynamics \eqref{eq:multi_layer_dyn} until convergence. We collect the fixed points $\r_{0,t}^{(p)}$. $\L^{(p)}$ is updated following a gradient ascent of \eqref{lstar}, while $\W^{(p)}$ and $\mathbf{b}^{(p)}$ follow a gradient descent of (\ref{cf}):
\begin{linenomath*}
\begin{align}\label{eq:fwd_update}
    \Delta \L^{(p)} &\propto \left(\r_{\beta,t}^{(p)}\r_{\beta,t}^{(p)\top} - \L^{(p)}\right),\nonumber\\
    \Delta \W^{(p)} &\propto \left(\r_{\beta,t}^{(p)}\r_{\beta,t}^{(p-1)\top} - \r_{0,t}^{(p)}\r_{0,t}^{(p-1)\top}\right), \nonumber \\
    \Delta \mathbf{b}^{(p)} &\propto \left(\r_{\beta,t}^{(p)} - \r_{0,t}^{(p)}\right).
\end{align}
\end{linenomath*}

In practice, learning rates can be chosen differently to achieve the best performance. A constant prefactor before $\L^{(p)}$  can be added to achieve numerical stability. The above CSM algorithm is summarized in Algorithm \ref{alg:CSM}.

\begin{algorithm}[h]
   \caption{Constrative Similarity Matching (CSM)}
   \label{alg:CSM}
\begin{algorithmic}
   \STATE {\bfseries Input:} Initial  $\{\W^{(p)}\}$, $\{\L^{(p)}\}$, $\{\mathbf{b}^{(p)}\}$,$\{\r^{(p)}\}$, $p=1,\ldots,P$
   \FOR{$t=1$ {\bfseries to} $T$}
   \STATE Run the nudged phase neural dynamics \eqref{eq:multi_layer_dyn} with $\beta \ne 0$ until convergence, collect the fixed point $\{\r_{\beta,t}^{(p}\}$
   \STATE Run the free phase dynamics \eqref{eq:multi_layer_dyn} with $\beta = 0$ until convergence, collect fixed point $\{\r_{0,t}^{(p)}\}$
   \STATE Update $\{\L^{(p)}\}$, $\{\W^{(p)}\}$ and $\{\mathbf{b}^{(p)}\}$ according to (\ref{eq:fwd_update}).
   \ENDFOR
\end{algorithmic}
\end{algorithm}

\subsubsection{Relation to Gradient Descent}

\red{The CSM algorithm can be related to gradient descent in the $\beta \rightarrow 0$ limit using similar arguments as in  \citep{scellier2017equilibrium}. To see this explicitly, we first simplify the notation by collecting all $\W^{(p)}$ and $\b^{(p)}$ parameters under one vector variable $\boldsymbol\theta$, denote all the lateral connection matrices defined in \eqref{lstar} by $\mathbf{L}^*$, and represent the fixed points of the network by $\bar{\mathbf{r}}$.  Now the energy function can be written as $E(\boldsymbol\theta, \beta; \L^*, \bar{\r})$, where $\L^*$
 and $\bar{\mathbf{r}}$ depend on $\boldsymbol \theta$ and $\beta$ implicitly. In the limit of small $\beta$, one can approximate the energy function to leading order by
\begin{linenomath*}
 \begin{align}
     E(\boldsymbol\theta, \beta; \L^*, \bar{\r}) \approx E(\boldsymbol\theta, 0) + \left.\left({\rm Tr}\left(\frac{\partial E}{\partial \L^*} \frac{\partial \L^*}{\partial \beta}\right) + \frac{\partial E}{\partial \bar{\mathbf{r}}}\cdot \frac{\partial \bar{\mathbf{r}}}{\partial \beta}+\frac{\partial E}{\partial \beta}\right)\right |_{\beta=0}\beta.
 \end{align}
\end{linenomath*}
Note that the maximization in \eqref{lstar} implies $\frac{\partial E}{\partial \L^*} = \boldsymbol 0$. If $\frac{\partial E}{\partial \bar{\mathbf{r}}}$ is also $\boldsymbol 0$, i.e. the minima of \eqref{multil} are not on the boundaries but at the interior of the feasible set,  then in the limit $\beta \rightarrow 0$, the gradient of the contrastive function is the gradient of the mean square error function with respect to $\boldsymbol\theta$:
\begin{linenomath*}
\begin{align}
    \lim_{\beta \to 0}\frac 1 {\beta}\frac{\partial J}{\partial \boldsymbol \theta} = \frac{\partial }{\partial \boldsymbol \theta} \left.\frac{\partial E}{\partial \beta}\right|_{\beta=0} =  \frac{\partial }{\partial \boldsymbol \theta}\,\frac 1T \sum_{t=1}^T\left\Vert \bar{\r}_t^{P}-\z_t^{l}\right\Vert_2^2.
\end{align}
\end{linenomath*}

It is important to note that while $\beta \to 0$ limit of CSM is related to gradient descent, this limit is not necessarily the best performing one (as also observed in \citep{scellier2017equilibrium} for EP) and $\beta$ is a hyperparameter to be tuned. In Appendix \ref{sec:beta-gamma}, we present simulations that confirm the existence of an optimal $\beta$ away from $\beta = 0$.}

\subsubsection{Relation to Other Energy Based Learning Algorithms}
The CSM algorithm is similar in spirit to other contrastive algorithms, such as CHL and EP. Like these algorithms, CSM performs two runs of the neural dynamics in a ``free'' and a ``nudged'' phase. However, there are important differences. One major difference is that in CSM, the contrastive function is minimized by the feedforward weights. The lateral weights take part in the maximization of a different minimax objective \eqref{lstar}. In CHL and EP, such minimization is done with respect to all the weights.

As a consequence of this difference,  CSM uses a different update for lateral weights than CHL and EP. This  anti-Hebbian update  is different in two ways: 1) It has the opposite sign, i.e. EP and CHL nudged/clamped phase lateral updates are Hebbian. 2) No update is applied in the free phase. As we will demonstrate in numerical simulations, our lateral update imposes a competition between different units in the same layer. When network activity is constrained to be nonnegative, such lateral interactions are inhibitory and sparsify neural activity.

\red{Analogs of two hyperparameters of our algorithm play special roles in EP and CHL. The $\beta \to \infty$ limit of EP corresponds to clamping the output to the desired value in the nudged phase \citep{scellier2017equilibrium}. Similarly, the $\beta \to \infty$ limit of CSM also corresponds to training with fully clamped output units.  We discussed the gradient descent interpretation of the $\beta\to 0$ limits of both algorithms above. CHL is equivalent to backpropagation when feedback strength, the analog of our $\gamma$ parameter, vanishes \citep{xie2003equivalence}. In CSM, $\gamma$ is a hyperparameter to be tuned, which we explore in Appendix F.}


\subsection{Introducing structured connectivity}

We can also generalize the nudged supervised similarity matching (Eq.\ref{eq:SM}) to derive a Hebbian/anti-Hebbian network with structured connectivity. Following  \cite{obeid2019structured}, we can modify any of the cross terms in the layer-wise similarity matching objective (Eq.\ref{eq:SM})
 by introducing synapse-specific structure constants. For example:
\begin{equation}
    -\frac{1}{T^2}\sum_i^{N^{(p)}}\sum_j^{N^{(p-1)}}\sum_t^T\sum_{t'}^Tr_{t,i}^{(p)}r_{t',i}^{(p)}r_{t,j}^{(p-1)}r_{t',j}^{(p-1)}s_{ij}^{W},
\end{equation}
where $N^{(p)}$ is the number of neurons in $p$-th layer, $s_{ij}^W \ge 0$ are constants that set the structure of feedforward weight matrix between $p$-th layer and $(p-1)$-th layer. In particular, setting them to zero removes the connection, without changing the interpretation of energy function \citep{obeid2019structured}. Similarly, we can introduce constants $s_{ij}^L$ to specify the structure of the lateral connections (Fig. \ref{fig: structured} A).  Using such structure constants, one can introduce many different architectures, some of which we experiment with below. We present a detailed explanation of these points in Appendix \ref{sec: appen_B}. 

\begin{figure}[t]
\centering
\includegraphics[width=1\linewidth]{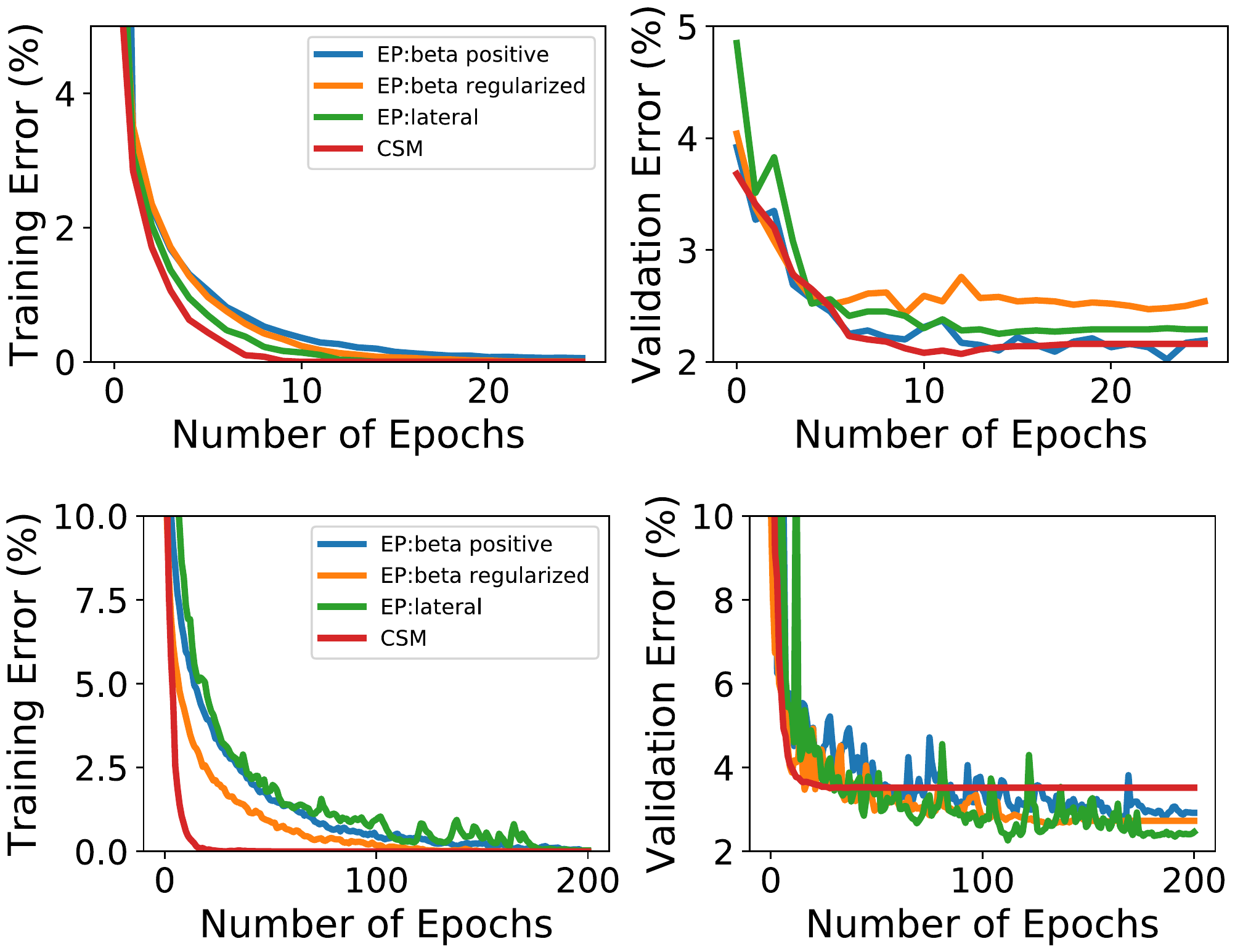}
\caption{Comparison of training (left) and validation (right) errors between CSM and EP algorithms for a network with one hidden layer (784-500-10, upper panels) and three hidden layers (784-500-500-500-10, lower panels) trained on the MNIST dataset.}
\label{fig:perf_net3}
\end{figure}

\section{Numerical Simulations}\label{sec:simulations}
In this section, we report the simulation results of the CSM algorithm on a supervised classification task using the MNIST dataset of handwritten digits \citep{lecun2010mnist} and the CIFAR-10 image dataset \citep{krizhevsky2009learning}. For our simulations, we used the Theano Deep Learning framework \citep{team2016theano} and modified the code released by \cite{scellier2017equilibrium}. The activation functions of the units were $f(x) = \min\{1,\max\{x,0\}\}$ and \red{$c^{(p)}=1/2 (1 - \delta_{pP})$}. \red{Following \cite{scellier2017equilibrium}, we used the persistent particle technique to tackle the long period of free phase relaxation. We stored the fixed points of hidden layers at the end of the free phase and used them to initialize the state of the network in the next epoch.}

\subsection{MNIST}
The inputs consist of gray-scale 28-by-28 pixel images, and each image is associated with a label ranging from $\{0,\cdots, 9\}$. We encoded the labels $\z^l$ as one-hot 10-dimensional vectors. We trained fully connected NNs with one and three hidden layers with lateral connections within each hidden layer. The performance of CSM algorithm was compared with several variants of EP algorithm: (1) EP: beta regularized, where the networks had no lateral connections and the sign of $\beta$ was randomized to act as a reqularizer as in \citep{scellier2017equilibrium} ; (2) EP: beta positive, where  the networks had no lateral connections and $\beta$ was a positive constant; (3)  EP: lateral, where networks had lateral connections and were trained with a positive constant $\beta$. In all the fully-connected network simulations for MNIST, the number of neurons in each hidden layer is 500. We attained 0\% training error and $2.16\%$ and $3.52\%$ validation errors with CSM, in the one and three hidden layer cases respectively. This is on par with the performance of the EP algorithm, which attains a validation error of $2.53\%$ and $2.73\%$ respectively for variant 1 and $2.18\%$ and $2.77\%$ for variant 2 (Fig.\ref{fig:perf_net3}). In the 3-layer case, a training error-dependent adaptive learning rate scheme (CSM-Adaptive) was used, wherein the learning rate for the lateral updates is successively decreased when the training error drops below certain thresholds (see Appendix \ref{sec:params_performance} for details).

\subsection{CIFAR-10}
\begin{figure}[h]
\centering
\includegraphics[width=\linewidth]{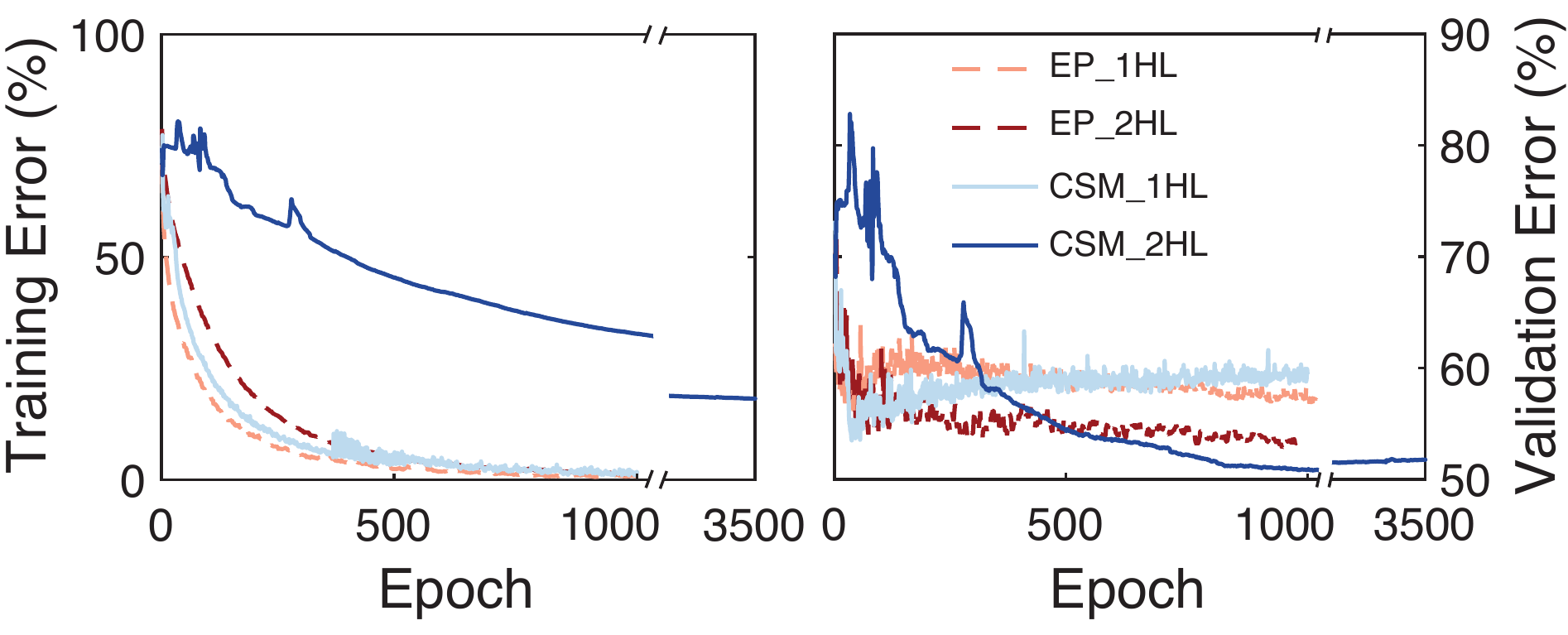}
\caption{Training (left) and validation (right) error curves for fully connected networks trained on CIFAR-10 dataset with CSM (solid) and EP (dashed) algorithms. The best fully connected CSM network attains slightly better validation accuracy than the best fully connected EP network.}
\label{fig:cifar10}
\end{figure}

CIFAR-10 is a more challenging dataset that contains 32-by-32 RGB images of objects belonging to ten classes of animals and vehicles. For fully connected networks, the performance of CSM was compared with EP (positive constant $\beta$). We obtain validation errors of 59.21\% and 51.76\% in the one and two hidden layer networks respectively in CSM, and validation errors of 57.60\% and 53.43\% in EP (Fig.\ref{fig:cifar10}). The mean and standard errors on the mean, of the last twenty validation errors, are reported here, in order to account for fluctuations about the mean. It is interesting to note that for both algorithms, deeper networks perform better for CIFAR-10, but not for MNIST. For both datasets, the best performing network trained with CSM achieves slightly better validation accuracy than the best performing network trained with EP. The errors corresponding to the fully connected networks for both algorithms and datasets are summarized in Table \ref{fc-table}. Here, CSM has been compared to the variant of EP with $\beta>0$.

\red{For CIFAR-10, the CSM algorithm with two hidden layers has a validation error around $51\%$ after 1000 epochs but was run for a total of 3584 epochs since the training error was still decreasing. The simulation does not reach zero training error, but starts plateauing at around 18\%, with a decrease of only 0.3\% for the last 100 epochs. The validation error does not decrease with the additional training beyond 1000 epochs. It is possible that better validation accuracy could be reached if better training errors were achieved e.g. by better performing learning rate schedules.}

\begin{table}[h]
\centering
\caption{Comparison of the training and validation errors of fully connected networks for EP (beta positive) and CSM. For both algorithms, the best performing networks correspond to two hidden layer networks for CIFAR-10 and one hidden layer networks for MNIST. Here, xHL means that the network has x hidden layers. For the CIFAR-10, CSM, 2HL simulation, errors at the end of 3584 epochs are reported. For the other CIFAR-10 simulations, errors at the end of 1000 epochs are reported.}
\label{fc-table}
\vskip 0.08in
\begin{small}
\begin{sc}
\renewcommand{\arraystretch}{1}
\begin{tabular}{l@{\hspace{0.5\tabcolsep}}c@{\hspace{1.5\tabcolsep}}c@{\hspace{1.0\tabcolsep}}c@{\hspace{1.0\tabcolsep}}cc@{\hspace{1.0\tabcolsep}}c@{\hspace{1.0\tabcolsep}}r}
\toprule
\multicolumn{3}{c}{MNIST} & \multicolumn{3}{c}{CIFAR-10} \\
\cmidrule(lr){1-3}\cmidrule(lr){4-6}
Rule & Train (\%) & Validate (\%)  & Rule & Train  (\%)& Validate (\%) \\ 
\midrule
CSM:1hl & 0.00 & 2.16 & CSM:1hl & 1.77 & $59.21\pm0.08$ \\
EP:1hl & 0.03 & 2.18 & EP:1hl & 0.76 & $57.60\pm 0.06$ \\
CSM:3hl & 0.00 & 3.52 & CSM:2hl & 17.96 & $51.76\pm0.002$ \\
EP:3hl & 0.00 & 2.77 & EP:2hl & 1.25 & $53.43\pm0.04$ \\
\bottomrule
\end{tabular}
\end{sc}
\end{small}
\vskip -0.1in
\end{table}

\subsection{Neuronal Representations}
\begin{figure}[t]
\centering
\includegraphics[width=\linewidth]{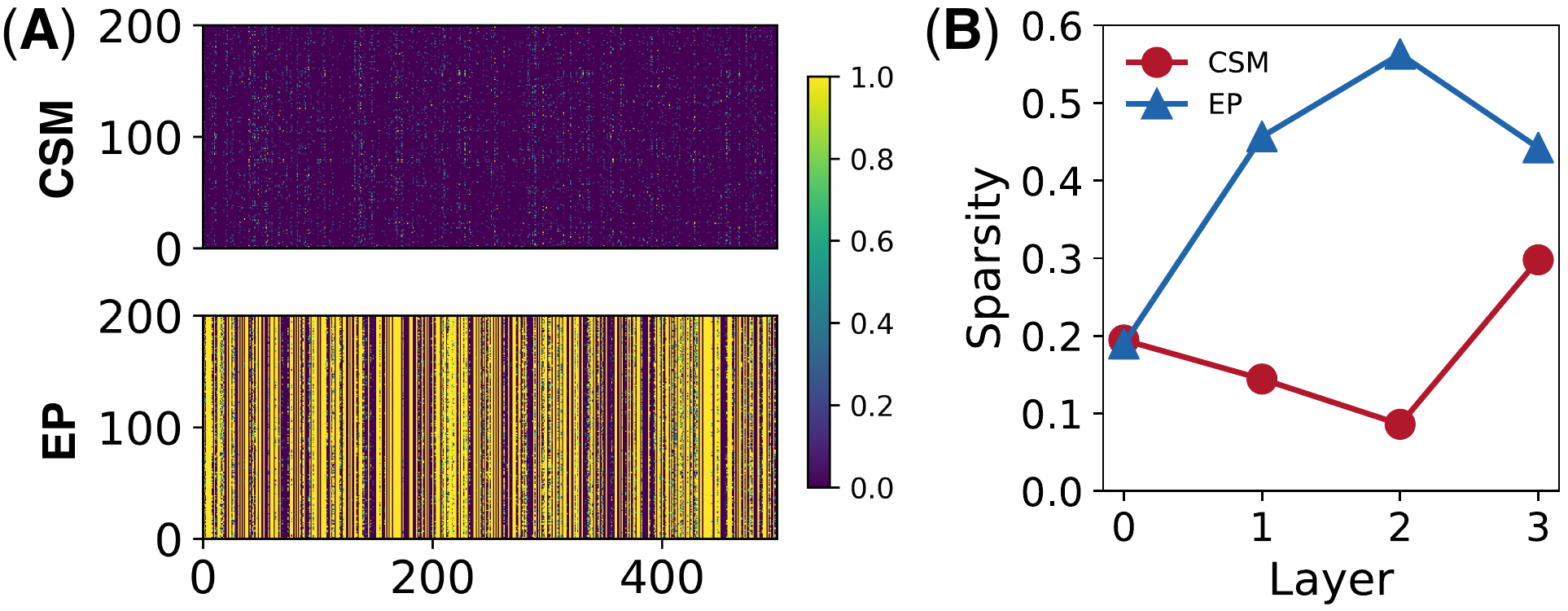}
\caption{Representations of neurons in NNs trained by CSM algorithm are much sparser than that of EP algorithm on MNIST dataset. (A) Heatmaps of representations at the second hidden layer, each row is the response of 500 neurons to a given digit image. Upper: CSM algorithm. Lower: EP algorithm. (B) Representation sparsity, defined as fraction of neurons whose activity are larger than a threshold (0.01), along different layers. Layer 0 is the input. The network has a 784-500-500-500-10 architecture.}
\label{fig:sparse_representation}
\end{figure}

    While CSM and EP perform similarly, their learned representations differ in sparseness (Fig.\ref{fig:sparse_representation}). \red{Due to the non-negativity of hidden unit activity and anti-Hebbian lateral updates, the CSM network ends up with inhibitory lateral connections, which enforce sparse response (Fig.\ref{fig:sparse_representation}). This can also be seen from the similarity matching objective \eqref{eq:SM}. Imagine there are only two inputs with a negative dot product, $\mathbf{x} \cdot \mathbf{x}' < 0$. The next layer will at least partially match this dot product, however, because the lowest value of $\mathbf{y}\cdot\mathbf{y}'$ is zero due to $\mathbf{y}, \mathbf{y}' \ge 0$,  $\mathbf{y}$ and $\mathbf{y}'$ will be forced to be orthogonal with non-overlapping sets of active neurons.} Sparse response is a general feature of cortical neurons \citep{olshausen2004sparse} and energy-efficient, making the representations learned by CSM more biologically relevant.

\subsection{Structured Networks}
\begin{figure}[h]
\centering
\includegraphics[width=0.6\linewidth]{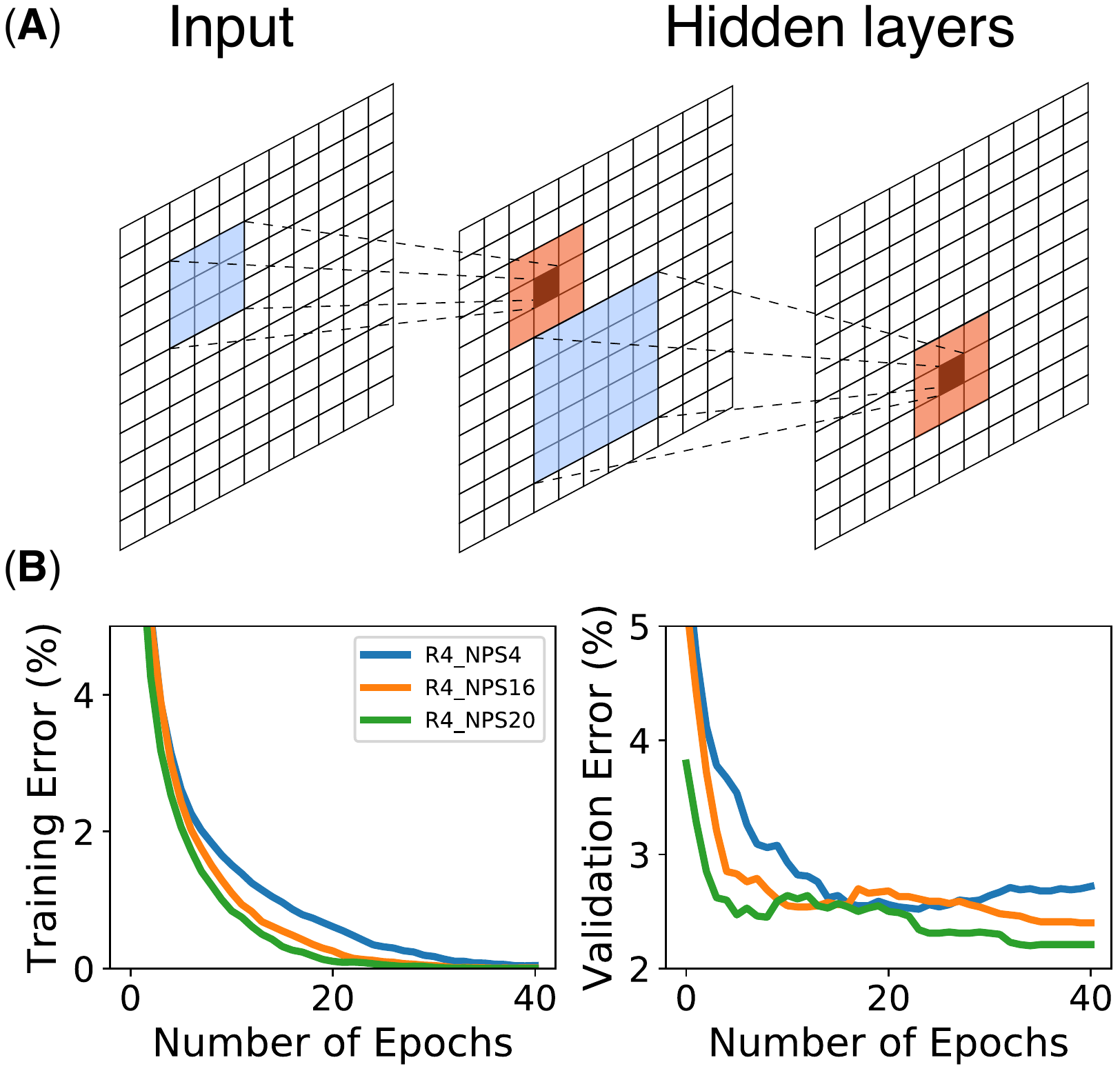}
\caption{(A) Sketch of structured connectivity in a deep neural network. Neurons live on a 2-d grid. Each neuron takes input from a small grid (blue shades) from the previous layer and a small grid of inhibition from its nearby neurons (orange shades).  (B) Training and validation curves of CSM with structured single hidden layer networks on MNIST dataset, with a receptive field of radius 4 and neurons per site 4, 16, and 20.}
\label{fig: structured}
\end{figure}

We also examined the performance of CSM in networks with structured connectivity. Every hidden layer can be constructed by first considering sites arranged on a two-dimensional grid. Each site only receives inputs from selected nearby sites controlled by the radius parameter (Fig.\ref{fig: structured}A). This setting resembles retinotopy \citep{kandel2000principles} in the visual cortex. Multiple neurons can be present at a single site controlled by the neurons per site (NPS) parameter. We consider lateral connections only between neurons sharing the same (x, y) coordinate.

For MNIST dataset, networks with structured connectivity trained with the CSM rule achieved $2.22\%$ validation error for a single hidden layer network with a radius of 4 and NPS of 20 (Fig. \ref{fig: structured}B) (See Appendix \ref{sec:params_performance} for details). For CIFAR-10 dataset, a one hidden layer structured network using CSM algorithm achieves 34\% training error and 49.5\% validation error after 250 epochs, which is a significant improvement compared to the fully connected one layer network. This structured network had a radius of 4 and NPS of 3. A two hidden layer structured network yielded a training error of 46.8\% and a validation error of 51.4\% after 200 epochs. Errors reported for the structured runs are the averages of five trials. The results for all fully connected and structured networks are reported in Appendix \ref{sec:params_performance} and \ref{sec:cifar_performance}.

\section{Discussion}\label{sec:discussion}
In this paper, we proposed a new solution to the credit assignment problem by generalizing the similarity matching principle to the supervised domain and proposed a biologically plausible supervised learning algorithm, the Contrastive Similarity Matching algorithm. In CSM, a supervision signal is introduced by minimizing the energy difference between a free phase and a nudged phase. CSM differs significantly from other energy-based algorithms in how the contrastive function is constructed. \red{We showed that when non-negativity constraint is imposed on neural activity}, the anti-Hebbian learning rule for the lateral connections makes the representations sparse and biologically relevant. We also derived the CSM algorithm for neural networks with structured connectivity.

The idea of using representational similarity for training neural networks has taken various forms in previous work. The similarity matching principle has recently been used to derive various biologically plausible unsupervised learning algorithms \citep{pehlevan2019neuroscience}, such as principal subspace projection \citep{pehlevan2015normative}, blind source separation \citep{pehlevan2017blind}, feature learning \citep{obeid2019structured}, and manifold learning \citep{sengupta2018manifold}. It has been used for semi-supervised classification \citep{genkin2019neural}.
 Similarity matching has also been used as part of a local cost function to train a deep convolutional network \citep{nokland2019training}, where instead of layer-wise similarity matching, each hidden layer aims to learn representations similar to the output layer. Representational similarity matrices derived from neurobiology data have recently been used to regularize CNNs trained for image classification. The resulting networks are more robust to noise and adversarial attacks \citep{li2019learning}. It would be interesting to study the robustness of neural networks trained by the CSM algorithm.

\red{Like other constrastive learning algorithms, CSM operates with two phases: free and nudged. Previous studies in contrastive learning provided various biologically possible implementations of such two phased learning. One proposal  is to introduce the teacher signal into the network through an oscillatory coupling with a period longer than the time scale of neural activity converging to a steady state. \cite{baldi1991contrastive} proposed that such oscillations might be related to rhythms in the brain. In more recent work,  \cite{scellier2017equilibrium} provided an implementation of EP also applicable to CSM with minor modifications. They proposed that synaptic update only happens in the nudged phase with weights continuously updating according to a differential anti-Hebbian rule as the neuron's state moves from the fixed point at free phase to the fixed point at the nudged phase.  Further, such differential rule can be related to spike time dependent plasticity \citep{xie2000spike,bengio2015stdp}. CSM can use the same mechanism for feedforward and feedback updates. Lateral connections need to be separately updated in the free phase.  The differential updating of synapses in different phases of the algorithm can be implemented by neuromodulatory gating of synaptic plasticity \citep{brzosko2019neuromodulation,bazzari2019neuromodulators}.} 

A practical issue of CSM and other energy-based algorithms such as EP and CHL is that the recurrent dynamics takes a long time to converge. Recently, a discrete-time version of EP has shown much faster training speed \citep{ernoult2019updates} and the application to the CSM could be an interesting future direction.

\section*{Acknowledgements}
We acknowledge support by NIH, the Intel Corporation through Intel Neuromorphic Research Community, and a Google Faculty Research Award. We thank Dina Obeid and Blake Bordelon for helpful discussions.

\appendix

\pagebreak

\section{Derivation of a Supervised Similarity Matching Neural Network}\label{sec:SSM}

The supervised similarity matching cost function \eqref{eq:targetFunLinear} is formulated in terms of the activities of units, but a statement about the architecture and the dynamics of the network has not been made. We will derive all these from the cost function, without prescribing them. To do so, we need to introduce variables that correspond to the synaptic weights in the network. As it turns out, these variables are dual to correlations between  unit activities \citep{pehlevan2018similarity}.

To see this explicitly, following the method of \citep{pehlevan2018similarity}, we expand the squares in Eq.\eqref{eq:targetFunLinear} and introduce new dual variables $\W_1 \in \mathbb{R}^{m\times n}$, $\W_2 \in \mathbb{R}^{k\times m}$ and $\mathbf{L}_1 \in \R^{m\times m}$ using the following identities:
\begin{linenomath*}
\begin{align}
         -\frac{1}{T^2}\sum_{t = 1}^T\sum_{t'=1}^T\y_t^{\top}\y_{t'}\x_t^{\top}\x_{t'} &= \min_{\W_1}-\frac{2}{T}\sum_{t=1}^T\y_t^{\top}\W_1\bf x_t +\Tr\W_1^{\top}\W_1, \nonumber\\
         \frac{1}{T^2}\sum_{t = 1}^T\sum_{t'=1}^T\y_t^{\top}\y_{t'}\y_t^{\top}\y_{t'} &= \max_{\L_1}\frac{2}{T}\sum_{t=1}^T\y_t^{\top}\L_1\y_t -\Tr\L_1^{\top}\bf L_1, \nonumber\\ 
         - \frac{1}{T^2}\sum_{t = 1}^T\sum_{t'=1}^T\y_t^{\top}\y_{t'}\z_t^{l\top}\z_{t'}^l &= \min_{\W_2}-\frac{2}{T}\sum_{t=1}^T\z_t^{l\top}\W_2\y_t +\Tr\W_2^{\top}\W_2.
\end{align}
\end{linenomath*}

Plugging these into Eq.\eqref{eq:targetFunLinear}, and changing orders of optimization, we arrive the following dual, minimax formulation of supervised similarity matching:
\begin{linenomath*}
\begin{equation}
    \min_{\bf W_1,\W_2} \max_{\L_1} \frac{1}{T}\sum_{t = 1}^T l_t(\W_1,\W_2,\L_1,\x_t,\z_t^l),
\end{equation}
where
\begin{align}\label{lt}
    l_t &:=\Tr\W_1^{\top}\W_1 - \Tr\L_1^{\top}\L_1 + \Tr\W_2^{\top}\W_2 \nonumber \\
    & + \min_{\y_t}2(-\y_t^{\top}\W_1\x_t + \y_t^{\top}\L_1\y_t - \y_t^{\top}\W_2^{\top}\z_t^l).
\end{align}
\end{linenomath*}

\begin{figure}[H]
\centering
\includegraphics[width=0.7\linewidth]{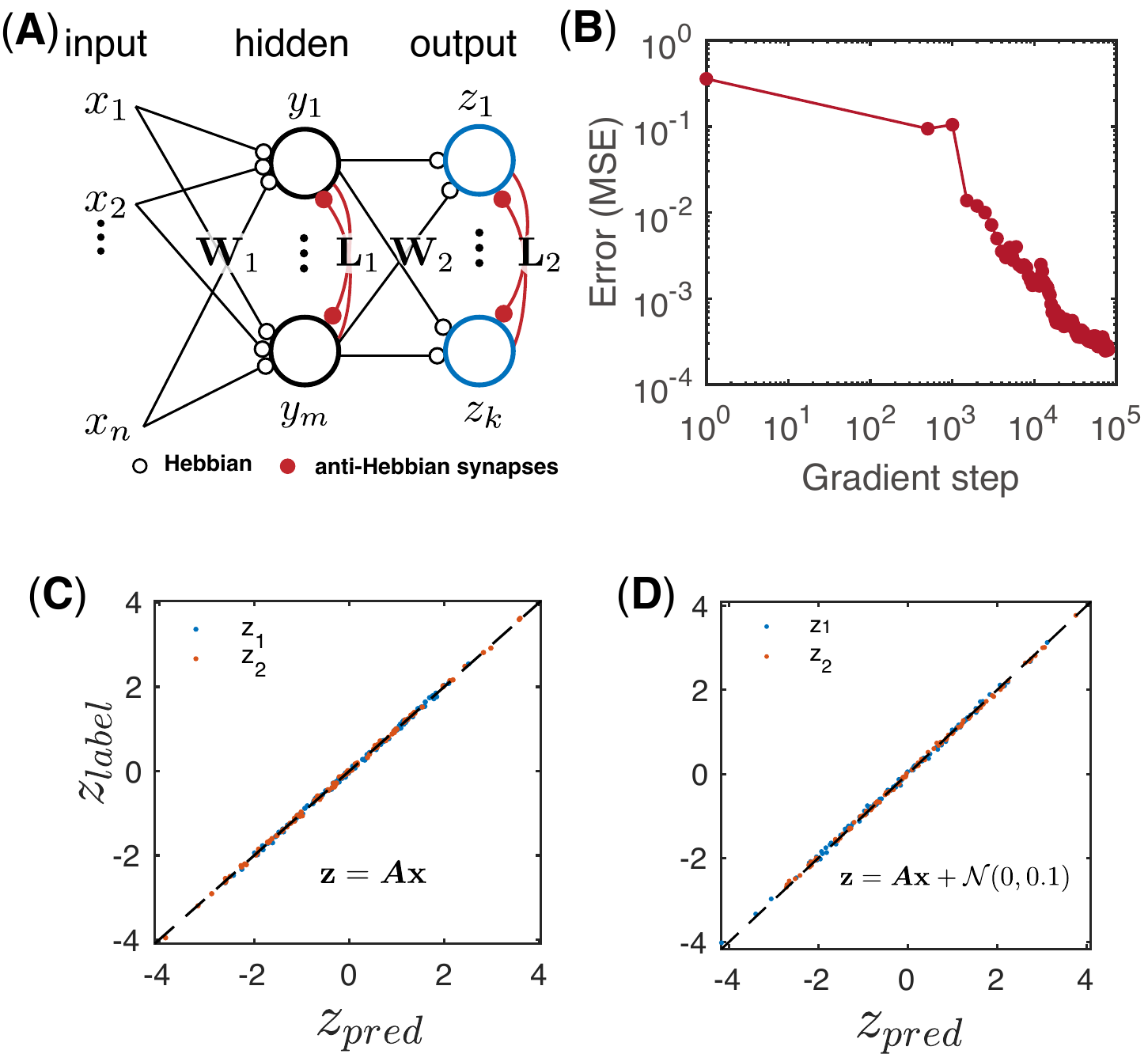}
\caption{A linear NN  with Hebbian/anti-Hebbian learning rules. (A) During the learning process, the output neurons (blue) are clamped at their desired states. After training, prediction for a new input $\x$ is given by the value of $\z$ at the fixed point of neural dynamics. (B) The network is trained on a linear task: $\z_t^l =\mathbf{A} \x_t$. Test error, defined as the mean square error between the network's prediction, $\z_t^p$, and the ground-truth value, $\z_t^l$, $1/T\sum_{t=1}^T||\z_t^p - \z_t^l||_F^2$, decreases with the gradient ascent-descent steps during learning. (C) Scatter plot of the predicted value versus the desired value (element-wise). (D) The algorithm learns the correct mapping between $\x$ and $\z$ even in the presence of small Gaussian noise. In these examples, $\x \in \R^5$, $\bm A\in \R^{2\times 5}$, elements of $\x$ and $\mathbf{A}$ are drawn from a uniform distribution in the range $[-1, 1]$, $\y \in \R^2$ and $\z \in \R^2$. In (C) and (D), 200 data points are shown. }
\label{fig:linearRegression}
\end{figure}

A stochastic optimization of the above objective can be mapped to a Hebbian/anti-Hebbian network following steps in \citep{pehlevan2018similarity}. For each training datum, $\{\x_t,\z^l_t\}$, a two-step procedure is performed. First, optimal $\y_t$ that minimizes $l_t$ is obtained by a gradient flow until convergence,
\begin{linenomath*}
\begin{equation}\label{eq:ydyn}
    \dot{\y} = \W_1\x_t - 2\L_1\y_t + \W_2^{\top}\z_t^l.
\end{equation}
\end{linenomath*}
We interpret this flow as the dynamics of a neural circuit with linear activation functions, where the dual variables $\W_1,\W_2$ and $\L_1$ are synaptic weight matrices (Fig. \ref{fig:linearRegression}A). In the second part of the algorithm, we update the synaptic weights by a gradient descent-ascent on \eqref{lt} with $\y_t$ fixed. This gives the following synaptic plasticity rules
\begin{linenomath*}
\begin{equation}\label{eq:updateRule}
\Delta\W_1  = \eta (\y_t\x_t^{\top} - \W_1), \quad \Delta\L_1  = \eta (\y_t\y_t^{\top} - \L_1), \quad \Delta\W_2  = \eta (\z_t^l\y_t^{\top} - \W_2).
\end{equation}
\end{linenomath*}
The learning rate $\eta$ of each matrix can be chosen differently to achieve best performance.

Overall, the network dynamics \eqref{eq:ydyn} and the update rules \eqref{eq:updateRule} map to a NN with one hidden layer, with the output layer clamped to the desired state. The updates of the feedforward weight are Hebbian, and updates of the lateral weight are anti-Hebbian (Fig. \ref{fig:linearRegression}A).

For prediction, the network takes an input data point, $\x_t$, and runs with unclamped output until convergence. We take the value of the $\z$ units at the fixed point as the network's prediction.

\red{Because the $\z$ units are not clamped during prediction and are dynamical variables, the correct outputs are not necessarily the fixed points of the network in the prediction phase.} To make sure that the network produces correct fixed points, at least for training data, we introduce the following step to the training procedure. We aim to construct a neural dynamics for the output layer in prediction phase such that its fixed point $\z$ corresponds to the desired output $\z^l$. Since the output layer receives input $\W_2\y$ from the previous layer, a decay term that depends on $\z$ is required to achieve stable fixed point at $\z=\z^l$. The simplest way is introducing lateral inhibition. And now the output layer has the following neural dynamics:
\begin{linenomath*}
\begin{equation}\label{eq:z_dyn}
    \dot{\z} = \W_2\y - \L_2\z,
\end{equation}
\end{linenomath*}
where the lateral connections $\L_2$ are learned such that the fixed point $\z^* \approx \z^l$. This is achieved by minimizing the following target function
\begin{linenomath*}
\begin{equation}\label{eq:lr_cost}
    \min_{\L_2} \frac{1}{T}\sum_{t=1}^T||\W_2\y_t - \L_2 \z_t^l||_2^2.
\end{equation}
\end{linenomath*}

Taking the derivative of the above target function with respect to $\L_2$ while keeping the other parameters and variables evaluated at the fixed point of neural dynamics, we get the following ``delta'' learning rule for $\L_2$:
\begin{linenomath*}
\begin{equation}\label{eq:updateL2}
   \Delta \L_2 = \eta(\W_2\y_t - \L_2\z_t^l)\z_t^{l\top}. 
\end{equation}
\end{linenomath*}

After learning, the NN makes a prediction about a new input $\x$ by running the neural dynamics of $\y$  and $\z$ \eqref{eq:ydyn} and \eqref{eq:z_dyn} until they converge to a fixed point. We take the value of $\z$ units at the fixed point as the prediction. As shown in Fig.\ref{fig:linearRegression} B-D, the linear network and weight update rule solve linear tasks efficiently.

Although the above procedure can be generalized to multi-layer and nonlinear networks, one has to address the issue of spurious fixed points of nonlinear dynamics for a given input $\x_t$.  The Contrastive Similarity Matching algorithm presented in the main text overcome this problem, which borrows ideas from energy-based learning algorithms such as Contrastive Hebbian Learning and Equilibrium Propagation.

\section{Supervised Deep Similarity Matching}
\label{sec:deep_SSM}
In this section, we follow \citep{obeid2019structured} to derive the minimax dual of deep similarity matching objective function. We start from rewriting the objective function \eqref{eq:SM} by expanding its first term and combining the same terms from adjacent layers, which gives
\begin{linenomath*}
\begin{equation}{\label{eq:SM_expand}}
\begin{split}
    &\min_{\substack{a_1\leq\r_t^{p}\leq a_2 \\ t = 1,\cdots, T \\ p = 1,\cdots,P}}\sum_{p=1}^P\frac{\gamma^{p-P}}{2T^2}\sum_{t=1}^T\sum_{t'=1}^T\left(\r_t^{(p)\top}\r_{t'}^{(p)}\r_t^{(p-1)\top}\r_{t'}^{(p-1)} -\frac{1+\gamma(1-\delta_{pP})}{2} c^{(p)}\r_t^{(p)\top}\r_{t'}^{(p)}\r_t^{(p)\top}\r_{t'}^{(p)}\right)  \\
    &+ \sum_{p = 1}^P\frac{2\gamma^{p-P}}{T}\sum_{t=1}^{T}\mathbf{F}(\r_t^{(p)})^{\top}\bm{1} + \frac \beta{T} \sum_{t=1}^T \left \Vert \r^{(P)}_t - \z^l_t\right\Vert_2^2,
\end{split}
\end{equation}
\end{linenomath*}

where $c^{(p)}$ is a parameter that change the relative importance of within-layer and between-layer similarity, we set it to be $1/2$ in our numerical simulations. Plug the following identities:
\begin{linenomath*}
\begin{align}
         -\frac{1}{T^2}\sum_{t = 1}^T\sum_{t'=1}^T\r_t^{(p)\top}\r_{t'}^{(p)}\r_t^{(p-1)\top}\r_{t'}^{(p-1)} &= \min_{\W^{(p)}}-\frac{2}{T}\sum_{t=1}^T\r_t^{(p)\top}\W^{(p)}\r_t^{(p-1)} +\Tr\W^{(p)\top}\W^{(p)}, \\
         \frac{1}{T^2}\sum_{t = 1}^T\sum_{t'=1}^T\r_t^{(p)\top}\r_{t'}^{(p)}\r_t^{(p)\top}\r_{t'}^{(p)} &= \max_{\L^{(p)}}\frac{2}{T}\sum_{t=1}^T\r_t^{(p)\top}\L^{(p)}\r_t^{(p)} -\Tr\L^{(p)\top}\L^{(p)},
\end{align}
\end{linenomath*}
in \eqref{eq:SM_expand} and exchange the optimization order of $\r_t^{(p)}$ and the weight matrices, we turn the target function \eqref{eq:SM} into the following minmax problem
\begin{linenomath*}
\begin{align}
\min_{\{\W^{(p)}\}}\max_{\{\L^{(p)}\}} &\frac 1T \sum_{t=1}^T l_t\left(\{\W^{(p)}\},\{\L^{(p)}\},\r^{(0)}_t,\z^l_t,\beta\right),
\end{align}
\end{linenomath*}
where we have defined an ``energy'' term (Eq.\ref{multil} in the main text).
The neural dynamics of each layer can be derived by following the gradient of $l_t$:
\begin{linenomath*}
\begin{align}
    \frac{d\bm{u}_t^{(p)}}{dt} &\propto -\frac{\partial l_t}{\partial \r_t^{(p)}}  =  2\gamma^{p-P}\left[-\bm{u}^{(p)} + \mathbf{b}_t^{(p)} + \W^{(p)}\r_t^{(p-1)} 
   + \gamma(1-\delta_{pP})\W^{(p+1)\top}\r_t^{(p+1)} \right. \nonumber\\
   & \left.- [1+\gamma (1-\delta_{pP})]c^{(p)}\L^{(p)}\r_t^{(p)} - 2\beta \delta_{pP}(\r_t^{(P)} - \z_t^l)\right],\nonumber\\
   \r_t^{(p)} &= \bm{f}(\bm{u}^{(p)}).
\end{align}
\end{linenomath*}
Define $\tau_p^{-1} = 2\gamma^{p-P}$, the above equation becomes Eq.\ref{eq:multi_layer_dyn} in the main text.

\section{Supervised Similarity Matching for Neural Networks with Structured Connectivity}\label{sec: appen_B}
In this section, we derive the supervised similarity matching algorithm for neural networks with structured connectivity. Structure can be introduced to the quartic terms in \eqref{eq:SM_expand}:
\begin{linenomath*}
\begin{align}
    &-\frac{1}{T^2}\sum_i^{N^{(p)}}\sum_j^{N^{(p-1)}}\sum_t^T\sum_{t'}^Tr_{t,i}^{(p)}r_{t',i}^{(p)}r_{t,j}^{(p-1)}r_{t',j}^{(p-1)}s_{ij}^{W,(p)}, \nonumber \\
   & -\frac{1}{T^2}\sum_i^{N^{(p)}}\sum_j^{N^{(p)}}\sum_t^T\sum_{t'}^Tr_{t,i}^{(p)}r_{t',i}^{(p)}r_{t,j}^{(p)}r_{t',j}^{(p)}s_{ij}^{L,(p)},
\end{align}
\end{linenomath*}
where $s_{ij}^{W,(p)}$ and $s_{ij}^{L,(p)}$ specify the feedforward connections of layer p with p-1 layer and lateral connections within layer p respectively. For example, setting them to be 0s eliminates all connections. Now we have the following deep structured similarity matching cost function for supervised learning:
\begin{linenomath*}
\begin{equation}{\label{eq:SM_structure}}
\begin{split}
    &\min_{\substack{a_1\leq\r_t^{p}\leq a_2 \\ t = 1,\cdots, T \\ p = 1,\cdots,P}}\sum_{p=1}^P\frac{\gamma^{p-P}}{2T^2}\sum_{t=1}^T\sum_{t'=1}^T\left(\r_t^{(p)\top}\r_{t'}^{(p)}\r_t^{(p-1)\top}\r_{t'}^{(p-1)}s_{ij}^{W,(p)} -\frac{1+\gamma(1-\delta_{pP})}{2} \r_t^{(p)\top}\r_{t'}^{(p)}\r_t^{(p)\top}\r_{t'}^{(p)}s_{ij}^{L,(p)}\right)  \\
    &+ \sum_{i = 1}^P\frac{2\gamma^{p-P}}{T}\sum_{t=1}^{T}\mathbf{F}(\r_t^{(p)})^{\top}\bm{1} + \frac \beta{T} \sum_{t=1}^T \left \Vert \r^{(P)}_t - \z^l_t\right\Vert_2^2.
\end{split}
\end{equation}
\end{linenomath*}
For each layer, we can define dual variables for $\W_{ij}^{(p)}$ and $\L_{ij}^{(p)}$ for interactions with positive constants, and define the following variables
\begin{linenomath*}
 \begin{equation}
    \bar{W}_{ij}^{(p)}=
    \begin{cases}
     W_{ij}^{(p)}, & s_{ij}^{W,(p)} \ne 0 \\
      0, & s_{ij}^{W,(p)} = 0
    \end{cases},\quad
    \bar{L}_{ij}^{(p)}=
    \begin{cases}
     L_{ij}^{(p)}, & s_{ij}^{L,(p)} \ne 0 \\
      0, & s_{ij}^{L,(p)} = 0
    \end{cases}
\end{equation}
\end{linenomath*}
Now we can rewrite \eqref{eq:SM_structure} as:
\begin{linenomath*}
\begin{align}
\min_{\{\bar{\W}^{(p)}\}}\max_{\{\bar{\L}^{(p)}\}} &\frac 1T \sum_{t=1}^T \bar{l}_t\left(\{\bar{\W}^{(p)}\},\{\bar{\L}^{(p)}\},\r^{(0)}_t,\z^l_t,\beta\right),
\end{align}
\end{linenomath*}
where
\begin{linenomath*}
\begin{align}\label{deep_structure}
\bar{l}_t : = &\min_{\substack{ a_1\leq \r_t^{(p)}  \leq a_2 \\ p = 1,\ldots,P}} \sum_{p=1}^P \gamma^{p-P} \Bigg\{\sum_{\substack{i,j \\ s_{ij}^{W,(p)} \ne 0}} W_{ij}^{(p)^2}  - \sum_{\substack{i,j \\ s_{ij}^{L,(p)} \ne 0}}\frac {1+\gamma(1- \delta_{pP})}{s_{ij}^{L,(p)}}L_{ij}^{(p)^2} + \nonumber\\
 &  \left[1+\gamma(1- \delta_{pP})\right]\r_t^{(p)}\L^{(p)}\r_t^{(p)}
 -2\r_t^{(p)\top}\W^{(p)}\r_t^{(p-1)} + 2 \mathbf{F}\left(\r_t^{(p)}\right)^{\top}\Bigg\} + \beta \left \Vert \r^{(P)}_t - \z^l_t\right\Vert_2^2.
\end{align}
\end{linenomath*}
The neural dynamics follows the gradient of \eqref{deep_structure}, which is
\begin{linenomath*}
\begin{align}
    \tau_p\frac{d\bm{u}_t^{(p)}}{dt} &= -\bm{u}^{(p)} + \mathbf{b}_t^{(p)} + \bar{\W}^{(p)}\r_t^{(p-1)}  + \gamma(1-\delta_{pP})\bar{\W}^{(p+1)\top}\r_t^{(p+1)} \nonumber\\ &\quad \, - [1+\gamma (1-\delta_{pP})]\bar{\L}^{(p)}\r_t^{(p)} 
    - \beta \delta_{pP}(\r_t^{(P)} - \z_t^l),\nonumber\\
   \r_t^{(p)} &= \bm{f}(\bm{u}^{(p)}), \quad p = 1, \cdots, P.
\end{align}
\end{linenomath*}
Local learning rules follow the gradient descent and ascent of \eqref{deep_structure}:
\begin{linenomath*}
\begin{align}
    \Delta W_{ij}^{(p)} &\propto \left(r_j^{(p)}r_i^{(p-1)} - \frac{W_{ij}^{(p)}}{s_{ij}^{W,(p)}}\right),\\
    \Delta L_{ij}^{(p)} &\propto \left(r_j^{(p)}r_i^{(p)} - \frac{L_{ij}^{(p)}}{s_{ij}^{L,(p)}}\right).
\end{align}
\end{linenomath*}

\section{Hyperparameters and Performance in Numerical Simulations with the MNIST Dataset}\label{sec:params_performance}
\subsection{One Hidden Layer}
Table \ref{1hlmn} reports the training and validation errors of three variants of the EP algorithm and the CSM algorithm for a single hidden layer network on MNIST. The models were trained until the training error dropped to $0\%$ or as close to $0\%$ as possible (as in the case of EP algorithm with $\beta>0$); errors reported herein correspond to errors obtained for specific runs and do not reflect ensemble averages. The training and validation errors below, and in subsequent subsections are reported at an epoch when the training error has dropped to $0\%$, or at the last epoch for the run (eg. for EP $\beta>0$). This epoch number is recorded in the last column. 

\begin{table}[h]
\caption{Comparison of the training and validation errors of different algorithms for one hidden layer NNs on MNIST data set}
\label{1hlmn}
\begin{center}
\begin{small}
\begin{sc}
\renewcommand{\arraystretch}{1}
\begin{tabular}{L{0.15\textwidth}C{0.25\textwidth}C{0.15\textwidth}C{0.15\textwidth}R{0.12\textwidth}}
\toprule
Algorithm & Learning Rate & Training Error (\%) & Validation Error (\%)  & No. epochs\\
\midrule
EP:$\pm\beta$    & $\alpha_W$ = 0.1, 0.05 & 0& 2.53 & 40 \\
EP:+$\beta$ & $\alpha_W$ = 0.5, 0.125 & 0.034 &2.18 & 100 \\
EP: lateral & $\alpha_W$ = 0.5, 0.25, $\alpha_L$ = 0.75 & 0 & 2.29 & 25 \\
CSM & $\alpha_W$ = 0.5, 0.375, $\alpha_L$ = 0.01 & 0 & 2.16 & 25 \\
\bottomrule
\end{tabular}
\end{sc}
\end{small}
\end{center}
\vskip -0.1in
\end{table}

\subsection{Three Hidden Layers}
In Table \ref{3hlmn}, the CSM algorithm employs a scheme with decaying learning rates. Specifically, the learning rates for lateral updates are divided by a factor of 5, 10, 50, and 100 when the training error dropped below $5\%, 1\%, 0.5\%$, and $0.1\%$ respectively.

\begin{table}[h]
\caption{Comparison of the training and validation errors of different algorithms for three hidden layer NNs on MNIST data set}
\label{3hlmn}
\begin{center}
\begin{small}
\begin{sc}
\renewcommand{\arraystretch}{1}
\begin{tabular}{L{0.12\textwidth}C{0.4\textwidth}C{0.12\textwidth}C{0.12\textwidth}R{0.12\textwidth}}
\toprule
Algorithm & Learning Rate & Training Error (\%) & Validation Error (\%)  & No. epochs\\
\midrule
EP: $\pm\beta$ & $\alpha_W$=0.128, 0.032, 0.008, 0.002 & 0 & 2.73 & 250 \\
\hline
EP: $+\beta$ & $\alpha_W $ =0.128, 0.032, 0.008, 0.002 & 0 & 2.77 & 250 \\
\hline
EP lateral & $\alpha_W $=0.128, 0.032, 0.008, 0.002; $\alpha_{L}=$ 0.192, 0.048, 0.012  & 0 & 2.4 & 250 \\
\hline
CSM & $\alpha_W=$0.5, 0.375, 0.281, 0.211; $\alpha_{L}=$0.75, 0.562, 0.422 & 0 & 4.82  & 250\\
\hline
CSM Adaptive &  $\alpha_W=$0.5, 0.375, 0.281, 0.211; $\alpha_{L}=$0.75, 0.562, 0.422  & 0 & 3.52 & 250 \\
\bottomrule
\end{tabular}
\end{sc}
\end{small}
\end{center}
\vskip -0.1in
\end{table}

\subsection{Structured Connectivity}
In this section, we explain the simulation for structured connectivity and report the results. Every hidden layer in these networks can be considered as multiple two dimensional grids stacked onto each other, with each grid containing neurons/units at periodically arranged sites. Each site only receives inputs from selected nearby sites. In this scheme, we consider lateral connections only between neurons sharing the same (x, y) coordinate, and the length and width of the grid are the same. In Table \ref{strmn}, `Full' refers to simulations where the input is the $28 \times 28$ MNIST input image and `Crop' refers to simulations in which the input image is a cropped $20 \times 20$ MNIST image. The first three, annotated by `Full',  correspond to the simulations reported in the main text. Errors are reported at the last epoch for the run. In networks with structural connectivity, additional hyperparameters are required to constrain the structure, which are enumerated below:
\begin{itemize}
\item Neurons-per-site (nps): The number of neurons placed at each site in a given hidden layer, i.e. the number of two dimensional grids stacked onto each other. The nps for the input is 1.
\item Stride: Spacing between adjacent sites, relative to the input channel. The stride of the input is always 1, i.e. sites are placed at (0, 0), (0, 1), (1, 0), so on, on the two dimensional grid. If the stride of the $l$-th layer is $s$, the nearest sites to the site at the origin will be $(0, s)$ and $(s,0)$. The stride increases deeper into the network. Specifying the stride also determines the dimension of the grid. A layer with stride s and nps $n$, will have $d \times d \times n$ units, where $d=28/s$ for the `Full' runs and $d=20/s$ for the `Crop' runs. The nps values and stride together assign coordinates to all the units in the network.
\item Radius: The radius of the circular two-dimensional region that all units in the previous layer must lie within in order to have non-zero weights to the current unit. Any units in the previous layer, lying outside the circle will not be connected to the unit.
\end{itemize}

\begin{table}[H]
\caption{Comparison of the training and validation errors of different algorithms for one hidden layer NNs with structured connectivity on MNIST data set}
\label{strmn}
\vskip 0.15in
\begin{center}
\begin{small}
\begin{sc}
\renewcommand{\arraystretch}{1}
\begin{tabular}{L{0.15\textwidth}C{0.25\textwidth}C{0.15\textwidth}C{0.15\textwidth}R{0.12\textwidth}}
\toprule
Algorithm & Learning Rate & Training Error (\%) & Validation Error (\%)  & No. epochs\\
\midrule
R4, NPS4, Full & $\alpha_{W}=$ 0.5, 0.375; $\alpha_{L}=$0.01 & 0.02 & 2.71 & 50 \\ 
\hline
R4, NPS16, Full & $\alpha_{W}=$ 0.5, 0.25; $\alpha_{L}=$0.75 & 0 & 2.41 & 49 \\ 
\hline
R4, NPS20, Full & $\alpha_{W}=$ 0.664, 0.577; $\alpha_{L}=$0.9 & 0 & 2.22 & 50 \\
\hline
R8, NPS80, Crop & $\alpha_{W}=$ 0.664, 0.577; $\alpha_{L}=$0.9 & 0.01 & 2.27 & 20 \\
\hline
R4, NPS4, Crop & $\alpha_{W}=$ 0.099, 0.065; $\alpha_{L}=$0.335 & 0.08 & 2.98 & 100 \\
\hline
R8, NPS4, Crop & $\alpha_{W}=$ 0.099, 0.065; $\alpha_{L}=$0.335 & 0 & 2.73 & 100 \\
\hline
R8, NPS20, Crop & $\alpha_{W}=$ 0.664, 0.577; $\alpha_{L}=$0.9 & 0 & 2.23 & 79 \\
\bottomrule
\end{tabular}
\end{sc}
\end{small}
\end{center}
\vskip -0.1in
\end{table}

\section{Hyperparameters and Performance in Numerical Simulations with the CIFAR-10 Dataset}\label{sec:cifar_performance}
Table \ref{allcif} records the training and validation errors obtained for the CSM and EP algorithms for fully connected networks, as well as for CSM with structured networks, on the CIFAR-10 dataset. The validation error column for fully connected runs reports the mean of the last twenty validation errors reported at the end of the training period as well as the standard error on the mean. For the structured runs, the training and validation errors reported are the average of the last epoch's reported errors from 5 trials  and the standard error on the means. This is done in order to account for fluctuations in the error during training. 
\begin{table}[h]
\caption{Comparison of the validation errors of different algorithms for different networks.}
\label{allcif}
\vskip 0.15in
\begin{center}
\begin{small}
\begin{sc}
\setlength{\tabcolsep}{3pt}
\renewcommand{\arraystretch}{1}
\begin{tabular}{L{0.2\textwidth}L{0.35\textwidth}C{0.1\textwidth}C{0.15\textwidth}R{0.1\textwidth}}
\toprule
Algorithm, Connectivity, No. Hidden Layers & Learning Rate & Train Error (\%) & Val Error (\%) & No. epochs \\
\midrule
CSM, FC, 1HL & $\alpha_W$ = 0.059, 0.017 & 1.77 & $59.21\pm 0.08$ & 1000 \\
    & $\alpha_L$ = 0.067    &     &  &  \\
\hline
CSM, FC, 2HL & $\alpha_W = 0.018, 7.51\times10^{-4}, 3.07\times10^{-5}$ & 17.96 & $51.76\pm 0.002$ & 3584 \\
 & $\alpha_L = 0.063, 2.59\times10^{-3}$ &     &     & \\
\hline
CSM, Str, 1HL & $\alpha_W$ = 0.050, 0.0375 & $34\pm3.7$ & $49.5\pm0.7$ & 250 \\
    & $\alpha_L$ = 0.01 &     &     & \\
\hline
CSM, Str, 2HL & $\alpha_W$ = 0.265, 0.073, 0.020 & $46.8\pm0.6$ & $51.4\pm0.7$ & 200 \\
    & $\alpha_L$ = 0.075, 0.020 &     &     & \\
\hline
EP, FC, 1HL & $\alpha_W$ = 0.014, 0.011 & 0.76 & $57.60\pm 0.06$ & 1000 \\
\hline
EP, FC, 2HL & $\alpha_W$ = 0.014, 0.011, 1.25 & 1.25 & $53.43\pm 0.04$ & 1000 \\
\bottomrule
\end{tabular}
\end{sc}
\end{small}
\end{center}
\vskip -0.1in
\end{table}

\section{Performance of CSM as a Function of Nudge Strength and Feedback Strength}\label{sec:beta-gamma}
\begin{figure}
\centering
\includegraphics[width=0.9\linewidth]{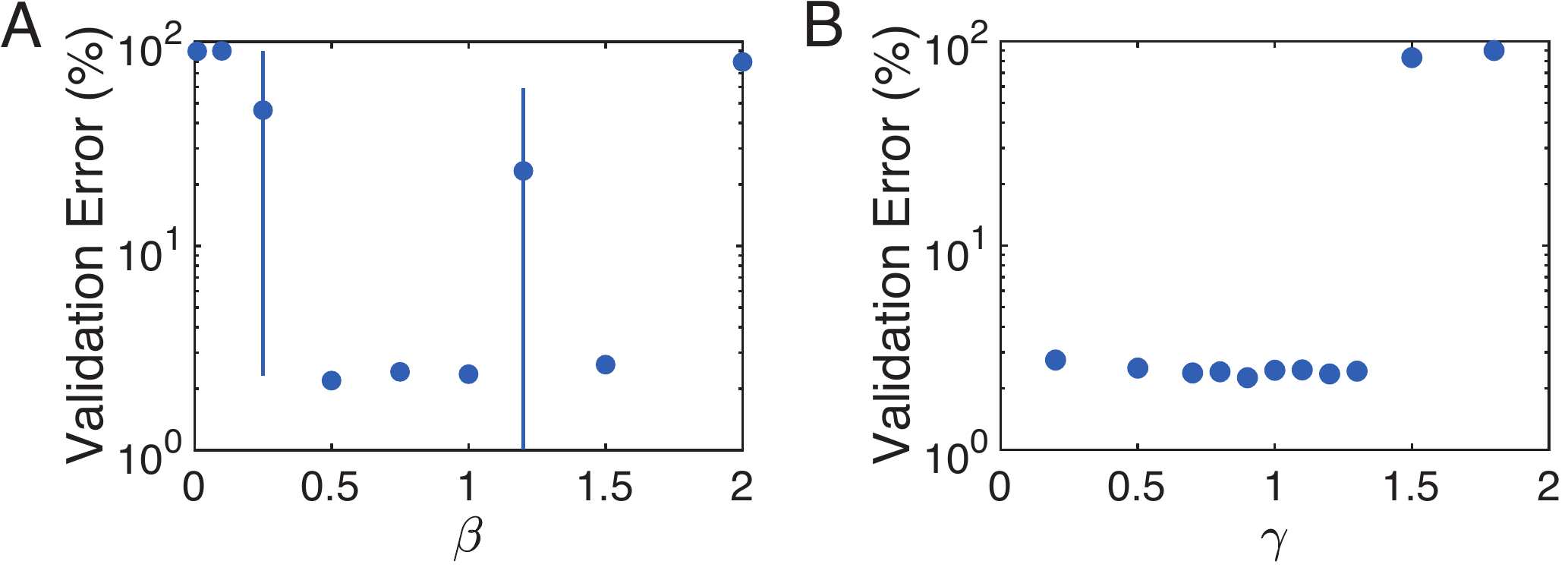}
\caption{Validation error of a single hidden layer network trained by CSM algorithm on MNIST dataset as a function of parameter $\beta$ (A)  and $\gamma$ (B). 4 trials were conducted for values for which the validation error was less than $3\%$. Dots indicate the mean validation error over trials, and errorbars indicate the standard deviation over trials.}
\label{fig:beta-gamma}
\end{figure}

In the nudged deep similarity matching objective \eqref{eq:SM}, $\beta$ controls the strength of nudging, while $\gamma$ specifies the strength of feedback input compared with the feedforward input. \citet{scellier2017equilibrium} used  $\beta=1$ in their simulations. In all the simulations reported here in the main text, we have set $\beta = 1, \gamma = 1$. In this section, we trained a single hidden layer network using CSM on MNIST, while systematically varying the value of $\beta$ and $\gamma$ and keeping other parameters fixed. The validation errors for these experiments are documented in Table \ref{beta-table} and \ref{gamma-table} and plotted in Fig.\ref{fig:beta-gamma}. We find that the network has optimal values for $\gamma$ less than 1.5, and $\beta$ in the range bounded by 0.5 and 1. At these values, the network is able to converge to low validation errors ($<3\%$).
 
\begin{table}[!htbp]
\caption{Validation errors at the end of the training period for a fully connected 1 hidden layer network trained on MNIST, with different $\beta$ values. For values that lay within the parameter range that converged to low ($< 3\%$) validation errors, 4 trials were run, and the mean, minimum and maximum errors over the trials have been reported. In all runs, $\gamma=1$.}
\label{beta-table}
\vskip 0.15in
\begin{center}
\begin{small}
\begin{sc}
\renewcommand{\arraystretch}{1}
\begin{tabular}{L{0.1\textwidth}C{0.25\textwidth}C{0.25\textwidth}R{0.25\textwidth}}
\toprule
$\beta$ value & Mean Validation Error (\%) & Minimum Validation Error (\%) & Maximum Validation Error (\%) \\
\midrule
0.01 & 89.70 & 89.70 & 89.70  \\
0.1 & 90.09 & 90.09 & 90.09 \\
0.25 & 46.20 & 2.28 & 90.09 \\
0.5 & 2.19 & 2.17 & 2.21 \\
0.75 & 2.42 & 2.22 & 2.51 \\
1.0 & 2.36 & 2.26 & 2.48 \\
1.2 & 23.30 & 2.36 & 85.91 \\
1.5 & 2.62 & 2.40 & 2.75 \\
2.0 & 79.55 & 79.55 & 79.55 \\
\bottomrule
\end{tabular}
\end{sc}
\end{small}
\end{center}
\vskip -0.1in
\end{table}

\begin{table}[!htbp]
\caption{Validation errors at the end of the training period for a fully connected 1 hidden layer network trained on MNIST, with different $\gamma$ values. For parameters that converged to low ($< 3\%$) validation errors, 4 trials were run, and the mean, minimum and maximum errors over the trials have been reported. In all runs, $\beta=1$.}
\label{gamma-table}
\vskip 0.15in
\begin{center}
\begin{small}
\begin{sc}
\renewcommand{\arraystretch}{1}
\begin{tabular}{L{0.1\textwidth}C{0.25\textwidth}C{0.25\textwidth}R{0.25\textwidth}}
\toprule
$\gamma$ value & Mean Validation Error (\%) & Minimum Validation Error (\%) & Maximum Validation Error (\%) \\
\midrule
0.2 & 2.75 & 2.64 & 2.85 \\
0.5 & 2.51 & 2.43 & 2.60 \\
0.7 & 2.38 & 2.24 & 2.47 \\
0.8 & 2.41 & 2.32 & 2.47 \\
0.9 & 2.26 & 2.21 & 2.31 \\
1.0 & 2.45 & 2.38 & 2.53 \\
1.1 & 2.46 & 2.37 & 2.63 \\
1.2 & 2.35 & 2.26 & 2.48 \\
1.3 & 2.43 & 2.28 & 2.54 \\
1.5 & 83.16 & 83.16 & 83.16 \\
1.8 & 90.09 & 90.09 & 90.09\\
\bottomrule
\end{tabular}
\end{sc}
\end{small}
\end{center}
\vskip -0.1in
\end{table}

\pagebreak

\end{document}